\documentclass[10pt,twocolumn,letterpaper]{article}
\usepackage{}

\usepackage{iccv}
\usepackage{times}
\usepackage{epsfig}
\usepackage{graphicx}
\usepackage{amsmath}
\usepackage{amssymb}
\usepackage{stfloats}

\usepackage{subfigure}
\usepackage{multirow}

\usepackage[pagebackref=true,breaklinks=true,letterpaper=true,colorlinks,bookmarks=false]{hyperref}

\iccvfinalcopy % *** Uncomment this line for the final submission

\def\iccvPaperID{2620} % *** Enter the ICCV Paper ID here

\ificcvfinal\fi
\begin{document}

\title{Learning Spatio-Temporal Representation with Pseudo-3D Residual Networks
\thanks{{\small This work was performed when Zhaofan Qiu was visiting Microsoft Research as a research intern. The codes and model of our P3D ResNet are publicly available at: https://github.com/ZhaofanQiu/pseudo-3d-residual-networks}}}
\author{Zhaofan Qiu $^{\dag}$, Ting Yao $^{\ddag}$, and Tao Mei $^{\ddag}$\\
$^{\dag}$ University of Science and Technology of China, Hefei, China\\
$^{\ddag}$ Microsoft Research, Beijing, China\\
{\tt\small zhaofanqiu@gmail.com, \{tiyao, tmei\}@microsoft.com}
}

\maketitle
\thispagestyle{empty}

%%%%%%%%% ABSTRACT
\begin{abstract}
Convolutional Neural Networks (CNN) have been regarded as a powerful class of models for image recognition problems. Nevertheless, it is not trivial when utilizing a CNN for learning spatio-temporal video representation. A few studies have shown that performing 3D convolutions is a rewarding approach to capture both spatial and temporal dimensions in videos. However, the development of a very deep 3D CNN from scratch results in expensive computational cost and memory demand. A valid question is why not recycle off-the-shelf 2D networks for a 3D CNN. In this paper, we devise multiple variants of bottleneck building blocks in a residual learning framework by simulating $3\times3\times3$ convolutions with $1\times3\times3$ convolutional filters on spatial domain (equivalent to 2D CNN) plus $3\times1\times1$ convolutions to construct temporal connections on adjacent feature maps in time. Furthermore, we propose a new architecture, named Pseudo-3D Residual Net (P3D ResNet), that exploits all the variants of blocks but composes each in different placement of ResNet, following the philosophy that enhancing structural diversity with going deep could improve the power of neural networks. Our P3D ResNet achieves clear improvements on Sports-1M video classification dataset against 3D CNN and frame-based 2D CNN by 5.3\% and 1.8\%, respectively. We further examine the generalization performance of video representation produced by our pre-trained P3D ResNet on five different benchmarks and three different tasks, demonstrating superior performances over several state-of-the-art techniques.
\end{abstract}

%%%%%%%%% BODY TEXT
\section{Introduction}
Today's digital contents are inherently multimedia: text, audio, image, video and so on. Images and videos, in particular, become a new way of communication between Internet users with the proliferation of sensor-rich mobile devices. This has encouraged the development of advanced techniques for a broad range of multimedia understanding applications. A fundamental progress that underlies the success of these technological advances is representation learning. Recently, the rise of Convolutional Neural Networks (CNN) convincingly demonstrates high capability of learning visual representation especially in image domain. For instance, an ensemble of residual nets \cite{he2015deep} achieves 3.57\% top-5 error on the ImageNet test set, which is even lower than 5.1\% of the reported human-level performance. Nevertheless, video is a temporal sequence of frames with large variations and complexities, resulting in difficulty in learning a powerful and generic spatio-temporal representation.

\begin{figure}
  \begin{minipage}{0.26\textwidth}
    \centering
    \includegraphics[width=1\textwidth]{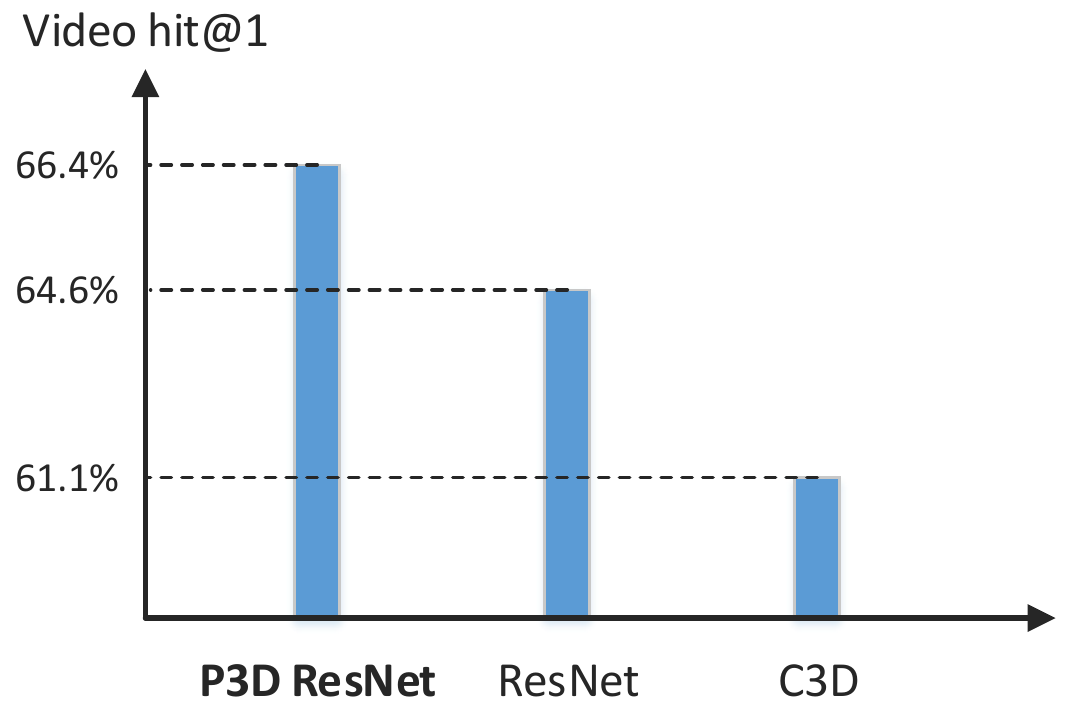}
  \end{minipage}
  \begin{minipage}{0.18\textwidth}
    \centering
    \scriptsize
    \begin{tabular}{l|c|c} \hline
      \multirow{2}{*}{\textbf{Method}} & \multirow{2}{*}{\textbf{Depth}}     & \textbf{Model} \\
                                       &                                     & \textbf{size} \\ \hline
      C3D                              & 11                                  & 321MB \\
      ResNet                           & 152                                 & 235MB \\ \hline
      \textbf{P3D ResNet}              & 199                                 & 261MB \\
      \hline
    \end{tabular}
  \end{minipage}
  \caption{\small Comparisons of different models on Sports-1M dataset in terms of accuracy, model size and the number of layers.}
     \label{fig:fig0}
  \vspace{-0.2in}
\end{figure}

One natural way to encode spatio-temporal information in videos is to extend the convolution kernels in CNN from 2D to 3D and train a brand new 3D CNN. As such, the networks have access not only the visual appearance present in each video frame, but also the temporal evolution across consecutive frames. While encouraging performances are reported in recent studies \cite{Ji:PAMI13,tran2015learning,varol2016long}, the training of 3D CNN is very computationally expensive and the model size also has a quadratic growth compared to 2D CNN. Take a widely adopted 11-layer 3D CNN, i.e., C3D \cite{tran2015learning} networks, as an example, the model size reaches 321MB which is even larger than that (235MB) of a 152-layer 2D ResNet (ResNet-152) \cite{he2015deep}, making it extremely difficult to train a very deep 3D CNN. More importantly, directly fine-tuning ResNet-152 with frames in Sports-1M dataset \cite{karpathy2014large} may achieve better accuracy than C3D trained on videos from scratch as shown in Figure \ref{fig:fig0}. Another alternative solution of producing spatio-temporal video representation is to utilize pooling strategy or Recurrent Neural Networks (RNN) over the representations of frames, which are often the activations of the last pooling layer or fully-connected layer in a 2D CNN. This category of approaches, however, only build temporal connections on the high-level features at the top layer while leaving the correlations in the low-level forms, e.g., edges at the bottom layers, not fully exploited.

We demonstrate in this paper that the above limitations can be mitigated by devising a family of bottleneck building blocks that leverages both spatial and temporal convolutional filters. Specifically, the key component in each block is a combination of one $1\times3\times3$ convolutional layer and one layer of $3\times1\times1$ convolutions in a parallel or cascaded fashion, that takes the place of a standard $3\times3\times3$ convolutional layer. As such, the model size is significantly reduced and the advantages of pre-learnt 2D CNN in image domain could also be fully leveraged by initializing the $1\times3\times3$ convolutional filters with $3\times3$ convolutions in 2D CNN. Furthermore, we propose a novel Pseudo-3D Residual Net (P3D ResNet) that composes each designed block in different placement throughout a whole ResNet-like architecture to enhance the structural diversity of the network. As a result, the temporal connections in our P3D ResNet are constructed at every level from bottom to top and the learnt video representations encapsulate information related to objects, scenes and actions in videos, making them generic for various video analysis tasks.

The main contribution of this work is the proposal of a family of bottleneck building blocks that simulates 3D convolutions in an economic and effective way. This also leads to the elegant view of how different blocks should be placed for learning very deep networks and a new P3D ResNet is presented for video representation learning. Through an extensive set of experiments, we demonstrate that our P3D ResNet outperforms several state-of-the-art models on five different benchmarks and three different tasks.

\section{Related Work}
We briefly group the methods for video representation learning into two categories: hand-crafted and deep learning-based methods.

Hand-crafted representation learning methods usually start by detecting spatio-temporal interest points and then describe these points with local representations. In this scheme, Space-Time Interest Points (STIP) \cite{laptev2005space}, Histogram of Gradient and Histogram of Optical Flow \cite{laptev2008learning}, 3D Histogram of Gradient \cite{klaser2008spatio} and SIFT-3D \cite{scovanner20073} are proposed by extending representations from image domain to measure the temporal dimension of 3D volumes. Recently, Wang \emph{et al.} propose dense trajectory features, which densely sample local patches from each frame at different scales and then track them in a dense optical flow field \cite{wang2013action}.

The most recent approaches for video representation learning are to devise deep architectures. Karparthy \emph{et al.} stack CNN-based frame-level representations in a fixed size of windows and then leverage spatio-temporal convolutions for learning video representation \cite{karpathy2014large}. In \cite{simonyan2014two}, the famous two-stream architecture is devised by applying two CNN architectures separately on visual frames and staked optical flows. This architecture is further extended by exploiting multi-granular structure \cite{li2016action,li2017learning,qiu2015msr}, convolutional fusion \cite{feichtenhofer2016convolutional}, key-volume mining \cite{zhu2016key} and temporal segment networks \cite{wang2016temporal} for video representation learning. In the work by Wang et al. \cite{wang2015action}, the local ConvNet responses over the spatio-temporal tubes centered at the trajectories are pooled as the video descriptors. Fisher Vector \cite{perronnin2010improving} is then used to encode these local descriptors to a global video representation.
Recently, the LSTM-RNN networks have been successfully employed for modeling temporal dynamics in videos. In \cite{jiang2017exploiting,yue2015beyond}, temporal pooling and stacked LSTM network are leveraged to combine frame-level (optical flow images) representation and discover long-term temporal relationships for learning a more robust video representation. Srivastava \emph{et al.} \cite{srivastava2015unsupervised} further formulate the video representation learning task as an autoencoder model based on the encoder and decoder LSTMs.

It can be observed that most aforementioned deep learning-based methods treat video as a frame/optical flow image sequence for video representation learning while leaving the temporal evolution across consecutive frames not fully exploited. To tackle this problem, 3D CNN proposed by Ji \emph{et al.} \cite{Ji:PAMI13} is one of the earlier works to directly learn the spatio-temporal representation of a short video clip. Later in \cite{tran2015learning}, Tran \emph{et al.} devise a widely adopted 11-layer 3D CNN (C3D) for learning video representation over 16-frame video clips in the context of large-scale supervised video datasets, and temporal convolutions across longer clips (100 frames) are further exploited in \cite{varol2016long}. However, the capacity of existing 3D CNN architectures is extremely limited with expensive computational cost and memory demand, making it hard to train a very deep 3D CNN. Our method is different that we not only propose the idea of simulating 3D convolutions with 2D spatial convolutions plus 1D temporal connections which is more economic, but also integrate this design into a deep residual learning framework for video representation learning.

\section{P3D Blocks and P3D ResNet}
In this section we firstly define the 3D convolutions for video representation learning which can be naturally decoupled into 2D spatial convolutions to encode spatial information and 1D temporal convolutional filters for temporal dimension. Then, a new family of bottleneck building blocks, namely Pseudo-3D (P3D), to leverage both spatial and temporal convolutional filters is devised in the residual learning framework. Finally, we develop a novel Pseudo-3D Residual Net (P3D ResNet) composing each P3D block at different placement in ResNet-like architecture and further compare its several variants through experimental studies in terms of both performance and time efficiency.

\subsection{3D Convolutions}\label{sec:3C}
Given a video clip with the size of $c \times l \times h \times w$ where $c$, $l$, $h$ and $w$ denotes the number of channels, clip length, height and width of each frame, respectively, the most natural way to encode the spatio-temporal information is to utilize 3D convolutions \cite{Ji:PAMI13,tran2015learning}. 3D convolutions simultaneously model the spatial information like 2D filters and construct temporal connections across frames. For simplicity, we denote the size of 3D convolutional filters as $d \times k \times k$ where $d$ is the temporal depth of kernel and $k$ is the kernel spatial size. Hence, suppose we have 3D convolutional filters with size of $3 \times 3 \times 3$, it can be naturally decoupled into $1 \times 3 \times 3$ convolutional filters equivalent to 2D CNN on spatial domain and $3 \times 1 \times 1$ convolutional filters like 1D CNN tailored to temporal domain. Such decoupled 3D convolutions can be regarded as a Pseudo 3D CNN, which not only reduces the model size significantly, but also enables the pre-training of 2D CNN from image data, endowing Pseudo 3D CNN more power of leveraging the knowledge of scenes and objects learnt from images.

\begin{figure}[!tb]
   \centering
   \subfigure[P3D-A]{
     \label{fig:fig1:a}
     \includegraphics[width=0.135\textwidth]{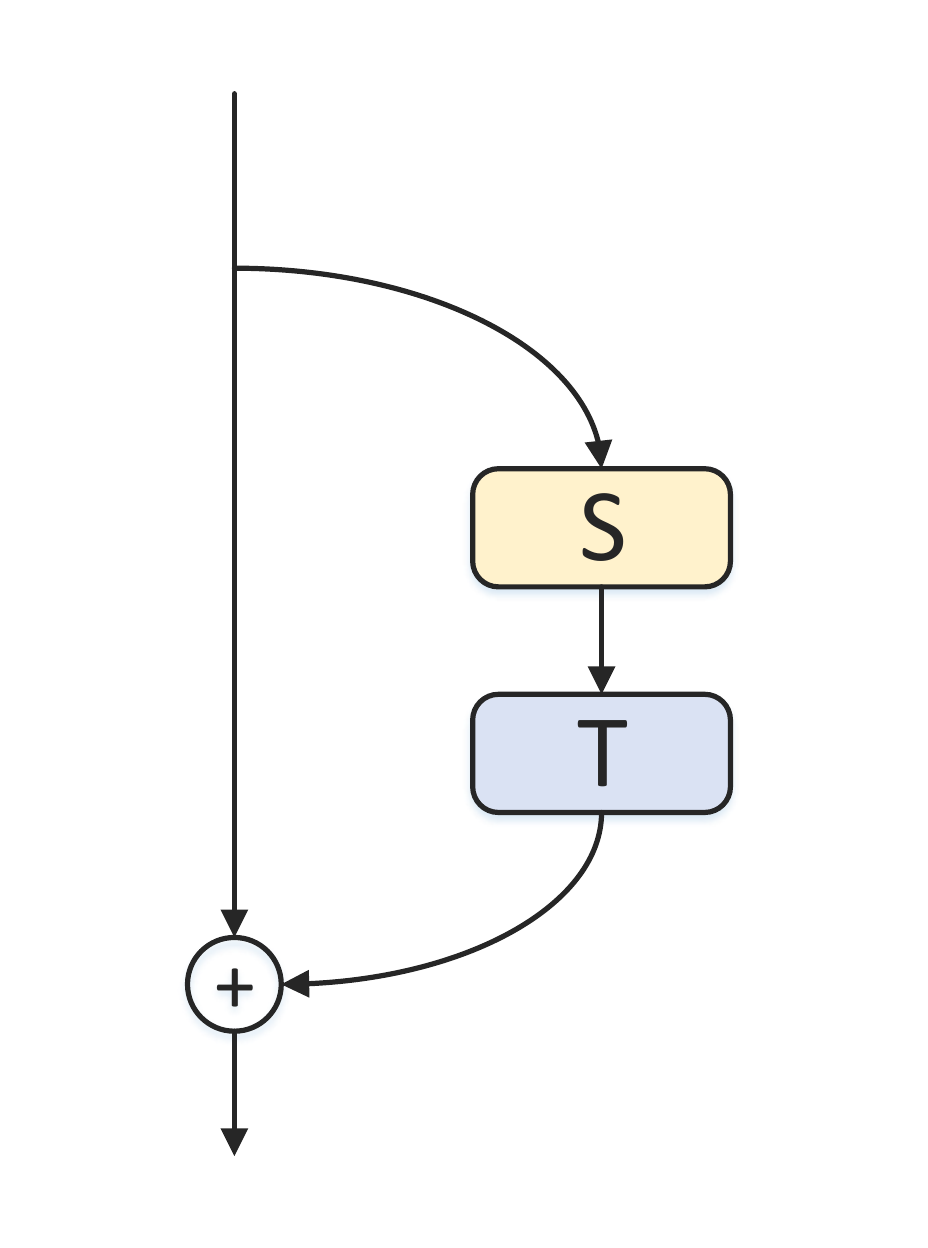}}
   \subfigure[P3D-B]{
     \label{fig:fig1:b}
     \includegraphics[width=0.135\textwidth]{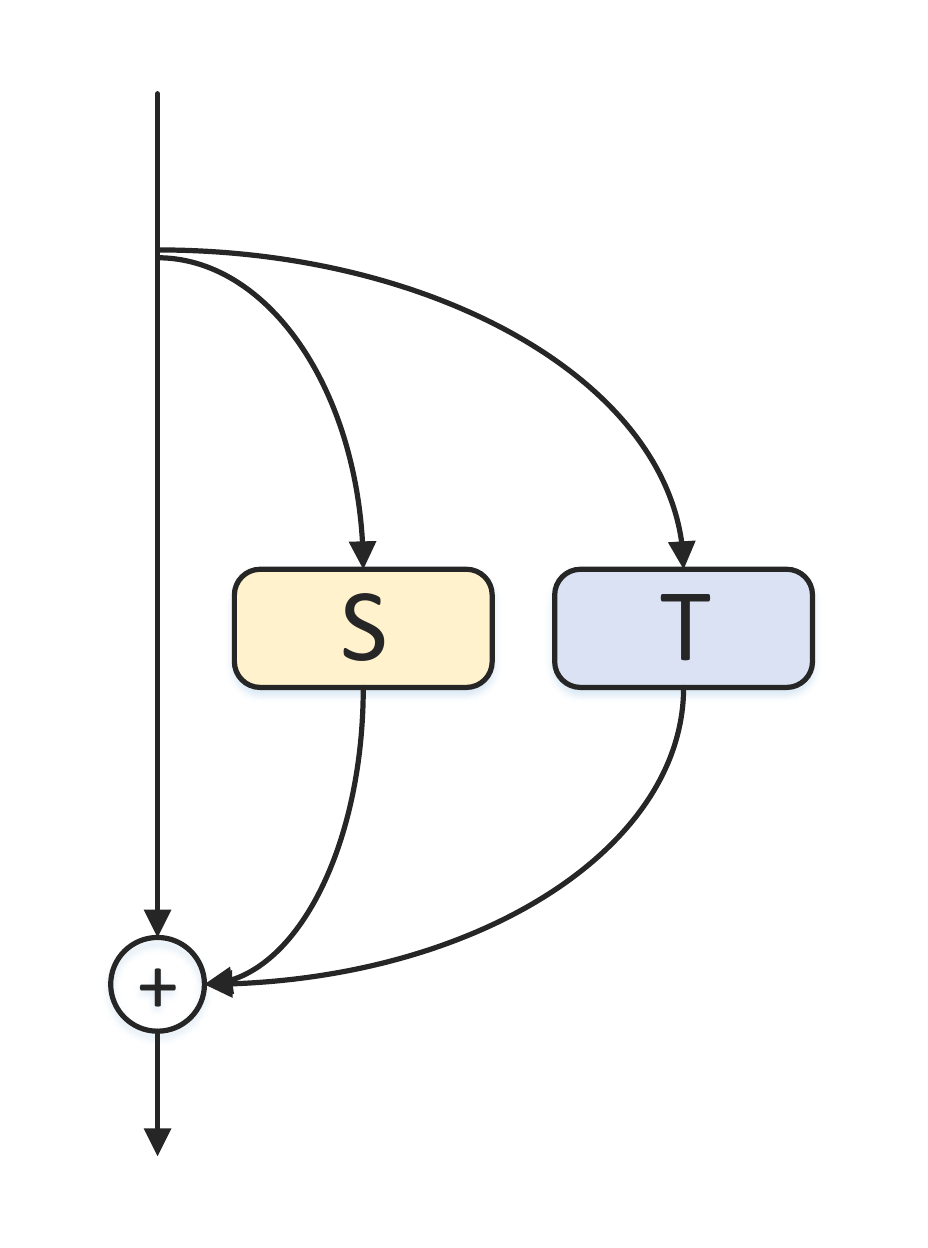}}
   \subfigure[P3D-C]{
     \label{fig:fig1:c}
     \includegraphics[width=0.135\textwidth]{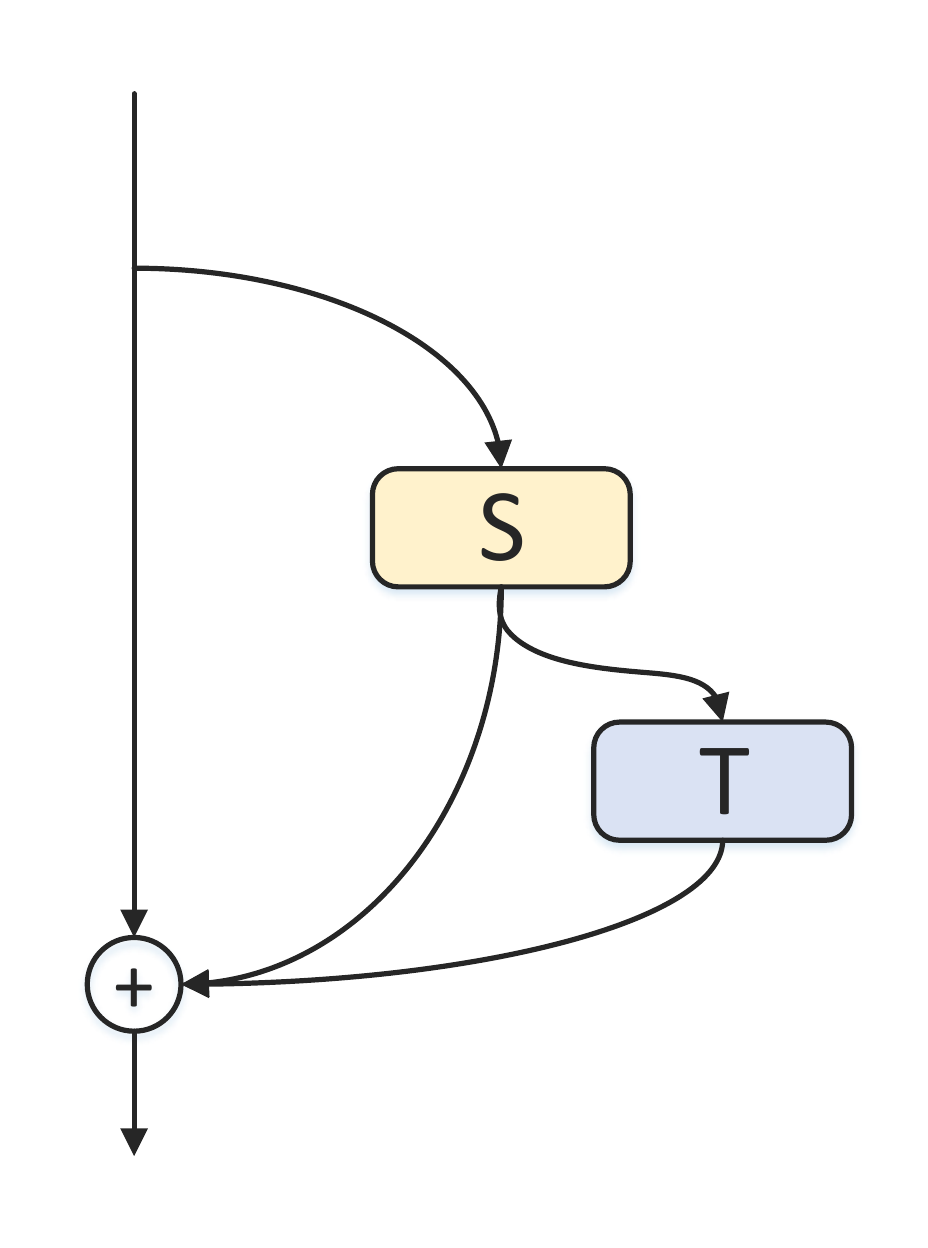}}
   \caption{\small Three designs of Pseudo-3D blocks.}
   \label{fig:fig1}
   \vspace{-0.16in}
\end{figure}

\subsection{Pseudo-3D Blocks}
Inspired by the recent successes of Residual Networks (ResNet) \cite{he2015deep} in numerous challenging image recognition tasks, we develop a new family of building modules named Pseudo-3D (P3D) blocks to replace 2D Residual Units in ResNet, pursuing spatio-temporal encoding in ResNet-like architectures for videos. Next, we will recall the basic design of Residual Units in ResNet, followed by presenting how to devise our P3D blocks. The bottleneck building architecture on each P3D block is finally elaborated.

\begin{figure*}[!tb]
   \centering
   \subfigure[Residual Unit \cite{he2015deep}]{
     \label{fig:fig2:a}
     \includegraphics[width=0.21\textwidth]{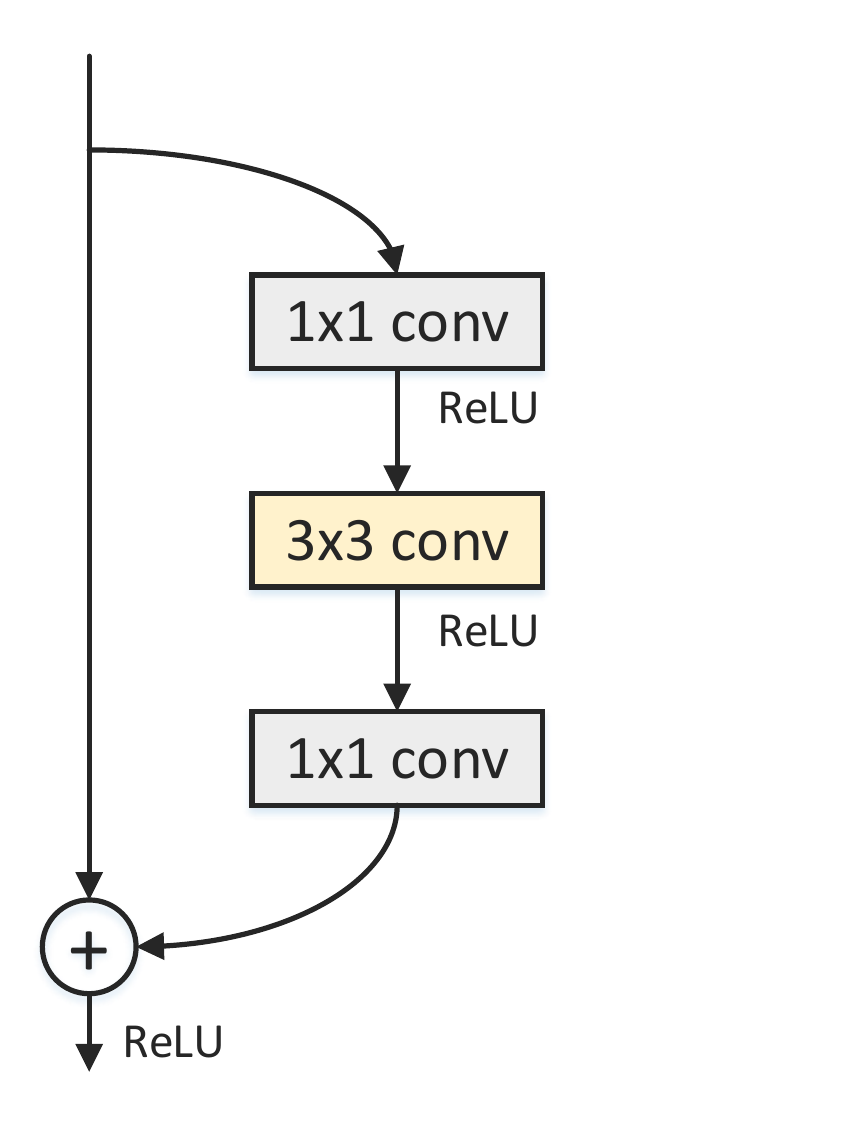}}
   \subfigure[P3D-A]{
     \label{fig:fig2:b}
     \includegraphics[width=0.21\textwidth]{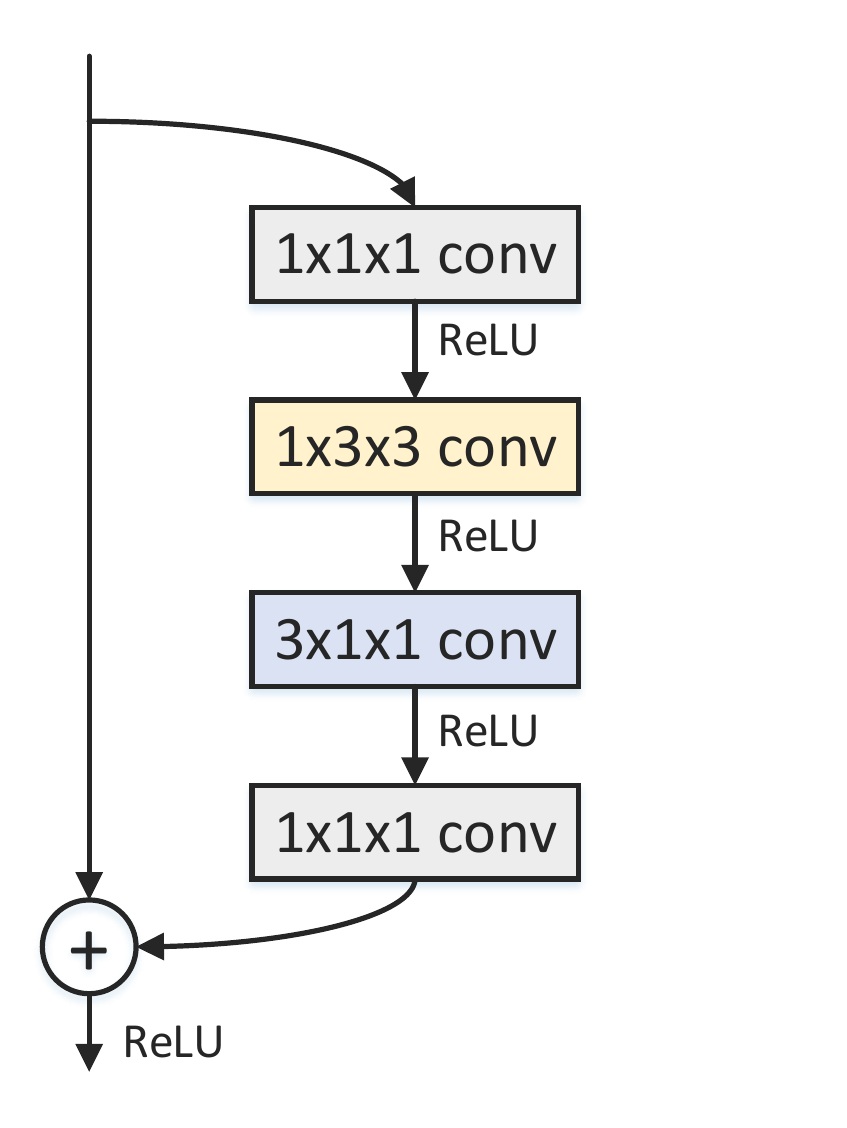}}
   \subfigure[P3D-B]{
     \label{fig:fig2:c}
     \includegraphics[width=0.21\textwidth]{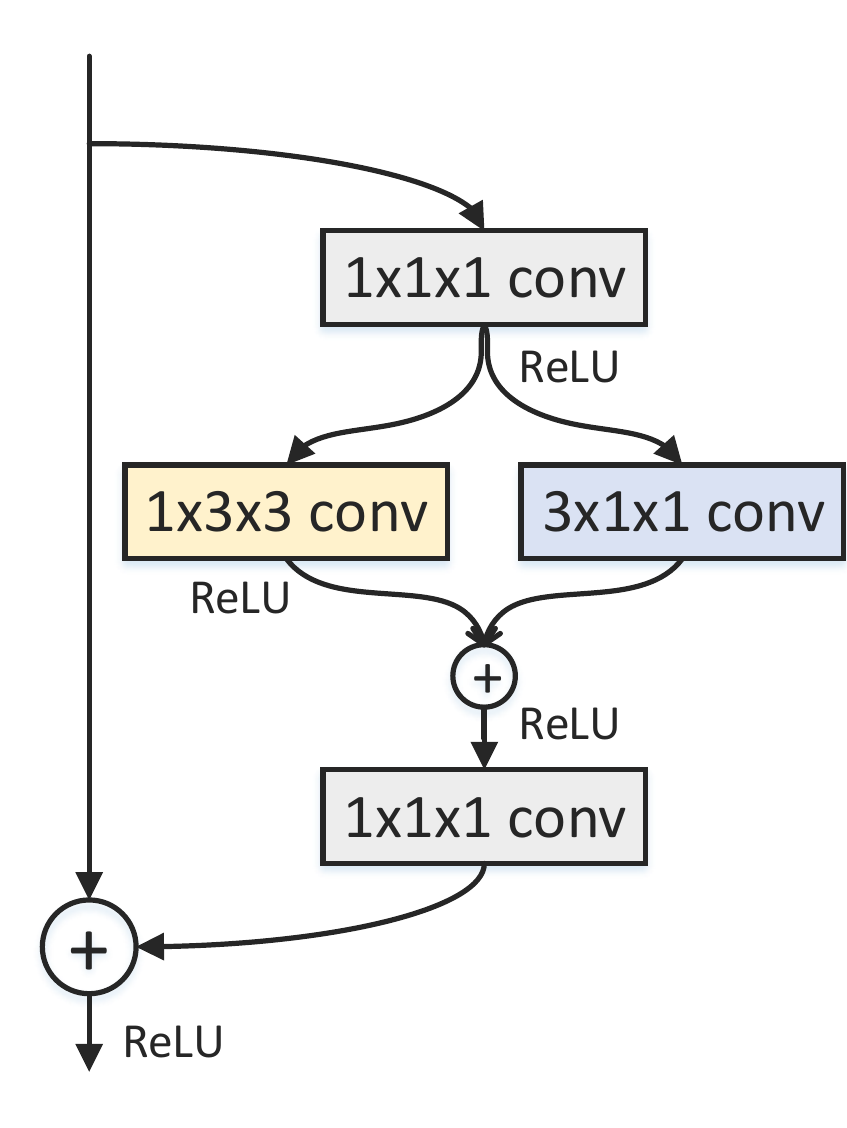}}
   \subfigure[P3D-C]{
     \label{fig:fig2:d}
     \includegraphics[width=0.21\textwidth]{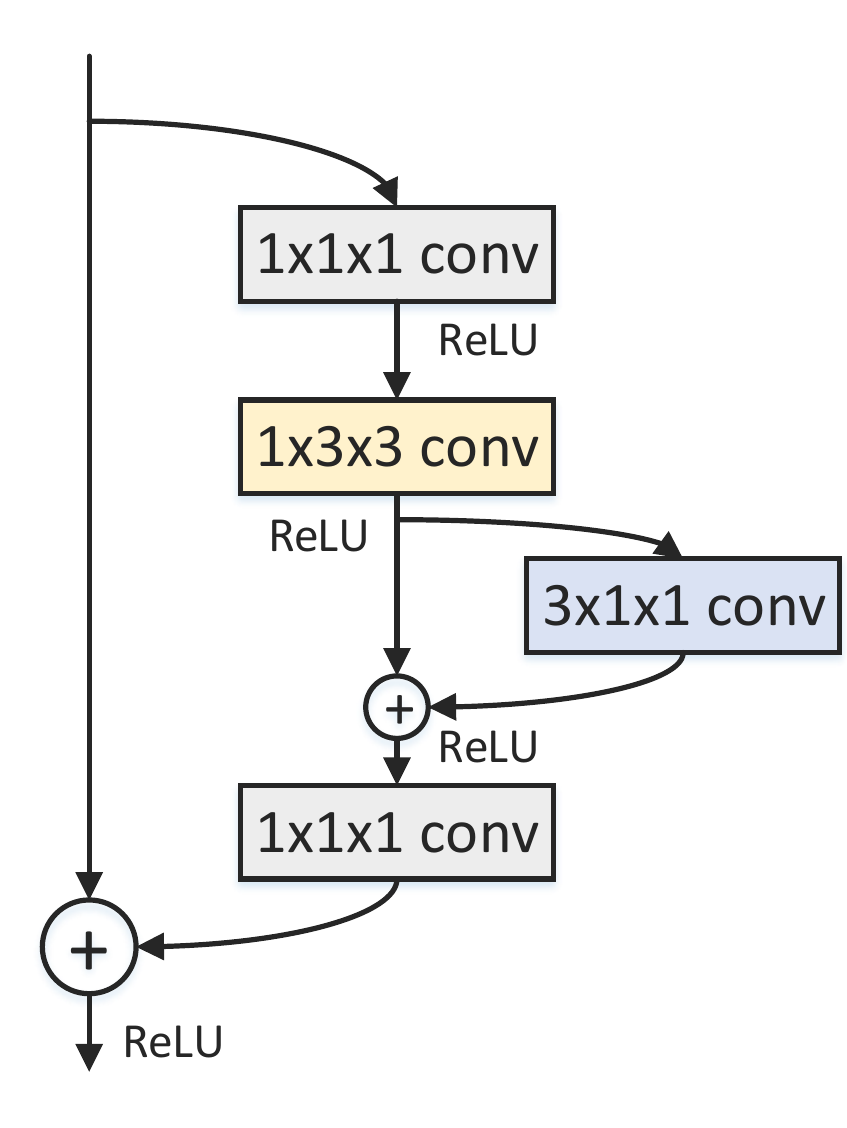}}
   \caption{\small Bottleneck building blocks of Residual Unit and our Pseudo-3D.}
   \label{fig:fig2}
   \vspace{-0.18in}
\end{figure*}

\textbf{Residual Units.}
ResNet consists of many staked Residual Units and each Residual Unit could be given~by
\begin{equation}\label{Eq:Eq1}\small
{{\bf{x}}_{t + 1}} = {\bf{h}}\left( {{{\bf{x}}_t}} \right) + {\bf{F}}\left( {{{\bf{x}}_t}} \right),
\end{equation}
where ${\bf{x}}_{t}$ and ${\bf{x}}_{t+1}$ denote the input and output of the $t$-th Residual Unit, ${\bf{h}}\left( {{{\bf{x}}_t}} \right)={{\bf{x}}_t}$ is an identity mapping and ${\bf{F}}$ is a non-linear residual function. Hence, Eq.(\ref{Eq:Eq1}) can be rewritten as
\begin{equation}\label{Eq:Eq2}\small
\left( {{\bf{I}} + {\bf{F}}} \right) \cdot {{\bf{x}}_t}={{\bf{x}}_t} + {\bf{F}} \cdot {{\bf{x}}_t} : = {{\bf{x}}_t} + {\bf{F}}\left( {{{\bf{x}}_t}} \right)={{\bf{x}}_{t + 1}},
\end{equation}
where ${\bf{F}} \cdot {{\bf{x}}_t}$ represents the result of performing residual function ${\bf{F}}$ over ${{\bf{x}}_t}$. The main idea of ResNet is to learn the additive residual function ${\bf{F}}$ with reference to the unit inputs ${\bf{x}}_{t}$ which is realized through a shortcut connection, instead of directly learning unreferenced non-linear functions.

\textbf{P3D Blocks design.}
To develop each 2D Residual Unit in ResNet into 3D architectures for encoding spatio-temporal video information, we modify the basic Residual Unit in ResNet following the principle of Pseudo 3D as introduced in Section \ref{sec:3C} and devise several Pseudo-3D Blocks. The modification is not straightforward for involvement of two design issues. The first issue is about whether the modules of 2D filters on spatial dimension (${\bf{S}}$) and 1D filters on temporal domain (${\bf{T}}$) should directly or indirectly influence each other. Direct influence within the two types of filters means that the output of spatial 2D filters is connected as the input to the temporal 1D filters (i.e., in a cascaded manner). Indirect influence between the two filters decouples the connection such that each kind of filters is on a different path of the network (i.e., in a parallel fashion). The second issue is whether the two kinds of filters should both directly influence the final output. As such, direct influence in this context denotes that the output of each type of filters should be directly connected to the final~output.

Based on the two design issues, we derive three different P3D blocks as depicted in Figure \ref{fig:fig1}, respectively, named as P3D-A to P3D-C. Detailed comparisons about their architectures are provided as following:

(1) P3D-A: The first design considers stacked architecture by making temporal 1D filters (${\bf{T}}$) follow spatial 2D filters (${\bf{S}}$) in a cascaded manner. Hence, the two kinds of filters can directly influence each other in the same path and only the temporal 1D filters are directly connected to the final output, which could be generally given by
\begin{equation}\label{Eq:Eq3}\small
\left({{\bf{I}} + {\bf{T}}\cdot{\bf{S}}}\right)\cdot {{\bf{x}}_t} := {{\bf{x}}_t} + {\bf{T}}\left( {{\bf{S}}\left( {{{\bf{x}}_t}} \right)} \right)={{\bf{x}}_{t + 1}}.
\end{equation}

%\begin{equation}\label{Eq:Eq3}\small
%\left({{\bf{I}} + {\bf{TS}}}\right)\cdot {{\bf{x}}_t} := {{\bf{x}}_t} + {\bf{TS}\left( {{{\bf{x}}_t}} \right)}={{\bf{x}}_{t + 1}}.
%\end{equation}

(2) P3D-B: The second design is similar to the first one except that indirect influence between two filters are adopted and both filters are at different pathways in a parallel fashion. Although there is no direct influence between ${\bf{S}}$ and ${\bf{T}}$, both of them are directly accumulated into the final output, which could be expressed as
\begin{equation}\label{Eq:Eq4}\small
\left({{\bf{I}} + {\bf{S}}+{\bf{T}}}\right)\cdot {{\bf{x}}_t} := {{\bf{x}}_t} + {\bf{S}}\left( {{{\bf{x}}_t}} \right) + {\bf{T}}\left( {{{\bf{x}}_t}} \right)={{\bf{x}}_{t + 1}}.
\end{equation}

(3) P3D-C: The last design is a compromise between P3D-A and P3D-B, by simultaneously building the direct influences among ${\bf{S}}$, ${\bf{T}}$ and the final output. Specifically, to enable the direct connection between ${\bf{S}}$ and final output based on the cascaded P3D-A architecture, we establish a shortcut connection from ${\bf{S}}$ to the final output, making the output ${{\bf{x}}_{t + 1}}$ as
\begin{equation}\label{Eq:Eq5}\small
\left({{\bf{I}} + {\bf{S}} + {\bf{T}}\cdot{\bf{S}}}\right)\cdot {{\bf{x}}_t} := {{\bf{x}}_t} + {\bf{S}}\left( {{{\bf{x}}_t}} \right) + {\bf{T}}\left( {{\bf{S}}\left( {{{\bf{x}}_t}} \right)} \right)={{\bf{x}}_{t + 1}}.
\end{equation}

\textbf{Bottleneck architectures.}
When specifying the architecture of 2D Residual Unit, the basic 2D block is modified with a bottleneck design for reducing the computation complexity. In particular, as shown in Figure \ref{fig:fig2:a}, instead of a single spatial 2D filters ($3 \times 3$ convolutions), the Residual Unit adopts a stack of $3$ layers including $1 \times 1$, $3 \times 3$, and $1 \times 1$ convolutions, where the first and last $1 \times 1$ convolutional layers are applied for reducing and restoring dimensions of input sample, respectively. Such bottleneck design makes the middle $3 \times 3$ convolutions as a bottleneck with smaller input and output dimensions. Thus, we follow this elegant recipe and utilize the bottleneck design to implement our proposed P3D blocks. Similar in spirit, for each P3D block which purely consists of one spatial 2D filters ($1 \times 3 \times 3$ convolutions) and one temporal 1D filters ($3 \times 1 \times 1$ convolutions), we additionally place two $1 \times 1 \times 1$ convolutions at both ends of the path, which are responsible for reducing and then increasing the dimensions. Accordingly, the dimensions of the input and output of both the spatial 2D and temporal 1D filters are reduced with this bottleneck design. The detailed bottleneck building architectures on all the three P3D blocks are illustrated in Figure \ref{fig:fig2:b} to \ref{fig:fig2:d}.

\subsection{Pseudo-3D ResNet}\label{ssec:PDR}
In order to verify the merit of the three P3D blocks, we first develop three P3D ResNet variants, i.e., P3D-A ResNet, P3D-B ResNet and P3D-C ResNet by replacing all the Residual Units in a 50-layer ResNet (ResNet-50) \cite{he2015deep} with one certain kind of P3D block, respectively. The comparisons of performance and time efficiency between the basic ResNet-50 and the three P3D ResNet variants are presented. Then, a complete version of P3D ResNet is proposed by mixing all the three P3D blocks from the viewpoint of structural diversity.

\begin{table}
\centering
\small
\caption{\small Comparisons of ResNet-50 and different Pseudo-3D ResNet variants in terms of model size, speed, and accuracy on UCF101 (split1). The speed is reported on one NVidia K40 GPU.}
\begin{tabular}{l|@{~~}c@{~~}|@{~~}c@{~~}|@{~~}c@{~~}} \hline
\textbf{Method}                                      & \textbf{Model size} & \textbf{Speed}  & \textbf{Accuracy} \\ \hline
ResNet-50                                            & 92MB                 & 15.0 frame/s    & 80.8\%           \\ \hline
P3D-A ResNet                                      & 98MB                 & 9.0 clip/s      & 83.7\%           \\
P3D-B ResNet                                      & 98MB                 & 8.8 clip/s      & 82.8\%           \\
P3D-C ResNet                                      & 98MB                 & 8.6 clip/s      & 83.0\%           \\ \hline
P3D ResNet                                        & 98MB                 & 8.8 clip/s      & 84.2\%           \\
\hline
\end{tabular}
\label{tab:s0m}
\vspace{-0.2in}
\end{table}

\textbf{Comparisons between P3D ResNet variants.} The comparisons are conducted on UCF101 \cite{UCF101} video action recognition dataset. Specifically, the architecture of ResNet-50 is fine-tuned on UCF101 videos. We set the input as $224\times224$ image which is randomly cropped from the resized $240\times320$ video frame. Moreover, following \cite{wang2016temporal}, we freeze the parameters of all Batch Normalization layers except for the first one and add an extra dropout layer with $0.9$ dropout rate to reduce the effect of over-fitting. After fine-tuning ResNet-50, the networks will predict one score for each frame and the video-level prediction score is calculated by averaging all frame-level scores. The architectures of three P3D ResNet variants are all initialized with ResNet-50 except for the additional temporal convolutions and are further fine-tuned on UCF101. For each P3D ResNet variant, the dimension of input video clip is set as $16\times160\times160$ which is randomly cropped from the resized non-overlapped 16-frame clip with the size of $16\times182\times242$. Each frame/clip is randomly horizontally flipped for augmentation. In the training, we set each mini-batch as 128 frames/clips, which are implemented with multiple GPUs in parallel. The network parameters are optimized by standard SGD and the initial learning rate is set as 0.001, which is divided by 10 after every 3K iterations. The training is stopped after 7.5K iterations.

\begin{table*}
\centering
\small
\caption{\small Comparisons in terms of pre-train data, clip length, Top-1 clip-level accuracy and Top-1\&5 video-level accuracy on Sports-1M.}
\begin{tabular}{l|c|c|c|c|c} \hline
\textbf{Method}                                     & \textbf{Pre-train Data} &\textbf{Clip Length} & \textbf{Clip hit@1}    & \textbf{Video hit@1}  & \textbf{Video hit@5} \\ \hline
Deep Video (Single Frame) \cite{karpathy2014large}  & ImageNet1K     & 1           & 41.1\%        & 59.3\%       & 77.7\%      \\
Deep Video (Slow Fusion) \cite{karpathy2014large}   & ImageNet1K     & 10          & 41.9\%        & 60.9\%       & 80.2\%      \\
Convolutional Pooling \cite{yue2015beyond}          & ImageNet1K     & 120         & 70.8\%        & 72.3\%       & 90.8\%      \\
C3D \cite{tran2015learning}                         & --             & 16          & 44.9\%        & 60.0\%       & 84.4\%      \\
C3D \cite{tran2015learning}                         & I380K          & 16          & 46.1\%        & 61.1\%       & 85.2\%      \\ \hline
ResNet-152 \cite{he2015deep}             & ImageNet1K     & 1           & 46.5\%        & 64.6\%       & 86.4\%      \\
P3D ResNet (ours)             & ImageNet1K     & 16          & 47.9\%        & 66.4\%       & 87.4\%      \\ \hline
\end{tabular}
\label{tab:s1m}
\vspace{-0.18in}
\end{table*}

Table \ref{tab:s0m} shows the performance and time efficiency of ResNet-50 and our Pseudo-3D ResNet variants on UCF101. Overall, all the three P3D ResNet variants (i.e., P3D-A ResNet, P3D-B ResNet and P3D-C ResNet) exhibit better performance than ResNet-50 with only a small increase in model size. The results basically indicate the advantage of exploring spatio-temporal information by our P3D blocks. Moreover, the speed of our P3D ResNet variants is very fast and could reach 8.6 $\sim$ 9.0 clips per second.

\begin{figure}[!tb]
   \centering {\includegraphics[width=0.47\textwidth]{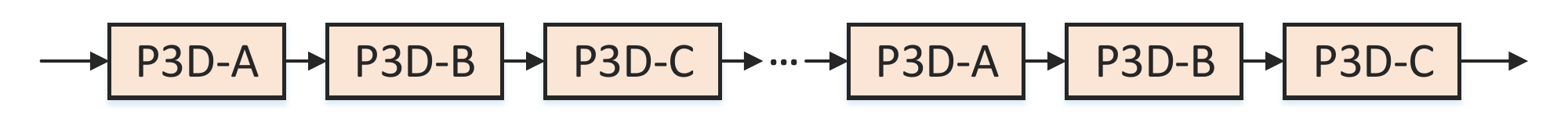}}
   \caption{\small P3D ResNet by interleaving P3D-A, P3D-B and P3D-C.}
   \label{fig:fig6}
   \vspace{-0.18in}
\end{figure}

\textbf{Mixing different P3D Blocks.}
Further inspired from the recent success of pursuing structural diversity in the design of very deep networks \cite{zhang2016polynet}, we devise a complete version of P3D ResNet by mixing different P3D blocks in the architecture to enhance structural diversity, as depicted in Figure \ref{fig:fig6}. Particularly, we replace Residual Units with a chain of our P3D blocks in the order P3D-A$\to$P3D-B$\to$P3D-C. Table \ref{tab:s0m} also details the performance and speed of the complete P3D ResNet. By additionally pursuing structural diversity, P3D ResNet makes the absolute improvement over P3D-A ResNet, P3D-B ResNet and P3D-C ResNet by 0.5\%, 1.4\% and 1.2\% in accuracy respectively, indicating that enhancing structural diversity with going deep could improve the power of neural networks.

\section{Spatio-Temporal Representation Learning}
We further validate the complete design of our P3D ResNet on a deeper 152-layer ResNet \cite{he2015deep} and then produce a generic spatio-temporal video representation. The learning of P3D ResNet here was conducted on Sports-1M dataset \cite{karpathy2014large}, which is one of the largest video classification benchmark. It roughly contains about 1.13 million videos annotated with 487 Sports labels. There are 1K-3K videos per label and approximately 5\% of the videos are with more than one label. Please also note that about 9.2\% video URLs were dead when we downloaded the videos. Hence, we conducted the experiments on the remaining 1.02 million videos and followed the official split, i.e., 70\%, 10\% and 20\% for training, validation and test set,~respectively.

\textbf{Network Training.} For efficient training on the large Sports-1M training set, we randomly select five 5-second short videos from each video in the set. During training, the settings of data augmentation and mini-batch are the same as those in Section \ref{ssec:PDR} except that the dropout rate is set as $0.1$. The learning rate is also initialized as 0.001, and divided by 10 after every 60K iterations. The optimization will be complete after 150K batches.

\textbf{Network Testing.} We evaluate the performance of the learnt P3D ResNet by measuring video/clip classification accuracy on the test set. Specifically, we randomly sample 20 clips from each video and adopt a single center crop per clip, which is propagated through the network to obtain a clip-level prediction score. The video-level score is computed by averaging all the clip-level scores of a~video.

We compare the following approaches for performance evaluation: (1) Deep Video (Single Frame) and (Slow Fusion) \cite{karpathy2014large}. The former performs a CNN which is similar to the architecture in \cite{krizhevsky2012imagenet} on one single frame from each clip to predict a clip-level score and fuses multiple frames in each clip with different temporal extent throughout the network to achieve the clip-level prediction. (2) Convolutional Pooling \cite{yue2015beyond} exploits max-pooling over the final convolutional layer of GoogleNet \cite{szegedy2015going} across each clip's frames. (3) C3D \cite{tran2015learning} utilizes 3D convolutions on a clip volume to model the temporal information and the whole architecture could be trained on Sports-1M dataset from scratch or fine-tuned from the pre-trained model on I380K internal dataset collected in \cite{tran2015learning}. (4) ResNet-152 \cite{he2015deep}. In this run, a 152-layer ResNet is fine-tuned and employed on one frame from each clip to produce a clip-level score.

\begin{figure*}[!tb]
   \centering {\includegraphics[width=0.93\textwidth]{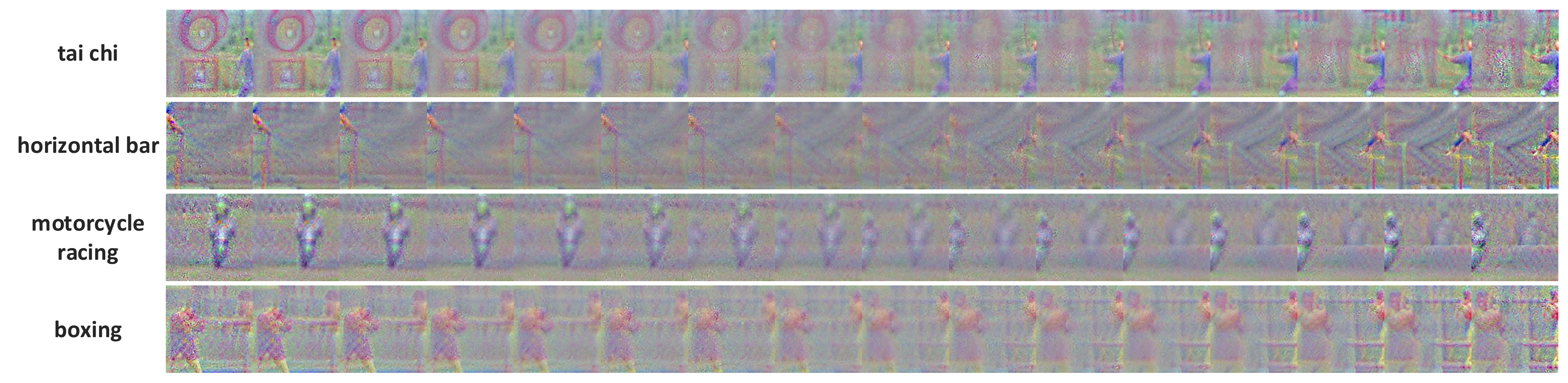}}
   \caption{\small Visualization of class knowledge inside P3D ResNet model by using DeepDraw \cite{deepdraw}. Four categories, i.e., tai chi, horizontal bar, motorcycle racing and boxing, are selected for visualization.}
   \label{fig:fig4}
   \vspace{-0.18in}
\end{figure*}

The performances and comparisons are summarized in Table \ref{tab:s1m}. Overall, our P3D ResNet leads to a performance boost against ResNet-152 (2D CNN) and C3D (3D CNN) by 1.8\% and 5.3\% in terms of top-1 video-level accuracy, respectively. The results basically indicate the advantage of exploring spatio-temporal information by decomposing 3D learning into 2D convolutions in spatial space and 1D operations in temporal dimension. As expected, Deep Video (Slow Fusion) fusing temporal information throughout the networks exhibits better performance than Deep Video (Single Frame) which exploits only one single frame. Though the three runs of Deep Video (Slow Fusion), Convolutional Pooling and our P3D ResNet all capitalizes on temporal fusion, they are fundamentally different in the way of performing temporal connections. The performance of Deep Video (Slow Fusion) is as a result of carrying out temporal convolutions on spatial convolutions to compute activations, while Convolutional Pooling is by simply max-pooling the outputs of final convolutional layer across temporal frames. As indicated by the results, our P3D ResNet employing different combinations of spatial and temporal convolutions improves Deep Video (Slow Fusion). This somewhat indicates that P3D ResNet is benefited from the principle of structural diversity in network design. It is also not surprise that the performances of P3D ResNet are still lower than Convolutional Pooling which performs temporal pooling on 120 frames' clips with frame rate of 1 fps, making the clip length over 120s. In contrast, we take 16 consecutive frames as a basic unit which only covers less than 0.5s but has strong spatio-temporal connections, making our P3D ResNet with better generalization~capability.

Figure \ref{fig:fig4} further visualizes the insights in the learnt P3D ResNet model. Following \cite{wang2016temporal}, we adopt DeepDraw toolbox \cite{deepdraw}, which conducts iterative gradient ascent on the input clip of white noises. During learning, it evaluates the model's violation of class label and back-propagates the gradients to modify the input clip. Thus, the final generated input clip could be regarded as the visualization of class knowledge inside P3D ResNet. We select four categories, i.e., tai chi, horizontal bar, motorcycle racing and boxing, for visualization. As illustrated in the figure, P3D ResNet model could capture both spatial visual patterns and temporal motion. Take the category of tai chi as an example, our model generates a video clip in which a person is displaying different poses, depicting the process of this action.

\textbf{P3D ResNet Representation.} After training our P3D ResNet architecture on Sports-1M dataset, the networks could be utilized as a generic representation extractor for any video analysis tasks. Given a video, we select 20 video clips and each clip is with 16-frame long. Each video clip is then input into the learnt P3D ResNet architecture and the 2,048 dimensional activations of pool5 layer are output as the representation of this clip. Finally, all the clip-level representations in a video are averaged to produce a 2,048 dimensional video representation. We refer to this representation as P3D ResNet representation in the following evaluations unless otherwise stated.

\section{Video Representation Evaluation}
Next, we evaluate our P3D ResNet video representation on three different tasks and five popular datasets, i.e., UCF101~\cite{UCF101}, ActivityNet~\cite{caba2015activitynet}, ASLAN \cite{KG:PAMI12}, YUPENN \cite{Der:CVPR12} and Dynamic Scene \cite{Shroff:CVPR10}. UCF101 and ActivityNet are two of the most popular video action recognition benchmarks. UCF101 consists of 13,320 videos from 101 action categories. Three training/test splits are provided by the dataset organisers and each split in UCF101 includes about 9.5K training and 3.7K test videos. The ActivityNet dataset is a large-scale video benchmark for human activity understanding. The latest released version of the dataset (v1.3) is exploited, which contains 19,994 videos from 200 activity categories. The 19,994 videos are divided into 10,024, 4,926 and 5,044 videos for training, validation and test set, respectively. It is also worth noting that the labels of test set are not publicly available and thus the performances on ActivityNet dataset are all reported on validation set.

ASLAN is a dataset on action similarity labeling task, which is to predict the similarity between videos. The dataset includes 3,697 videos from 432 action categories. We follow the strategy of 10-fold cross validation with the official data splits on this set. Furthermore, YUPENN and Dynamic Scene are two sets for the scenario of scene recognition. In between, YUPENN is comprised of 14 scene categories each containing 30 videos. Dynamic Scene consists of 13 scene classes with 10 videos per class. The training and test procedures on both datasets follow the standard leave-one-video-out protocol.

\begin{table}
\centering
\small
\caption{\small Performance comparisons with the state-of-the-art methods on UCF101 (3 splits). TSN: Temporal Segment Networks \cite{wang2016temporal}; TDD: Trajectory-pooled Deep-convolutional Descriptor \cite{wang2015action}; IDT: Improved Dense Trajectory \cite{wang2013action}. We group the approaches into three categories, i.e., end-to-end CNN architectures which are fine-tuned on UCF101 at the top, CNN-based video representation extractors with linear SVM classifier in the middle and approaches fused with IDT at the bottom. For the methods in the first direction, we report the performance of only taking frames and frames plus optical flow (in brackets) as inputs, respectively.}
\begin{tabular}{l|c} \hline
\textbf{Method}                                             & \textbf{Accuracy} \\ \hline
\multicolumn{2}{l}{End-to-end CNN architecture with fine-tuning} \\ \hline
Two-stream ConvNet \cite{simonyan2014two}                   & 73.0\%~~(88.0\%) \\
Factorized ST-ConvNet \cite{sun2015human}                   & 71.3\%~~(88.1\%) \\
Two-stream + LSTM \cite{yue2015beyond}                      & 82.6\%~~(88.6\%) \\
Two-stream fusion \cite{feichtenhofer2016convolutional}     & 82.6\%~~(92.5\%) \\
Long-term temporal ConvNet \cite{varol2016long}             & 82.4\%~~(91.7\%) \\
Key-volume mining CNN \cite{zhu2016key}                     & 84.5\%~~(93.1\%) \\
ST-ResNet \cite{feichtenhofer2016spatiotemporal}            & 82.2\%~~(93.4\%) \\
TSN \cite{wang2016temporal}                                 & 85.7\%~~(94.0\%) \\ \hline
\multicolumn{2}{l}{CNN-based representation extractor + linear SVM} \\ \hline
C3D \cite{tran2015learning}                                 & 82.3\%          \\
ResNet-152                                                   & 83.5\%          \\
\textbf{P3D ResNet}                                      & \textbf{88.6\%}          \\ \hline
\multicolumn{2}{l}{Method fusion with IDT} \\ \hline
IDT \cite{wang2013action}                                   & 85.9\%          \\
C3D + IDT \cite{tran2015learning}                           & 90.4\%          \\
TDD + IDT \cite{wang2015action}                             & 91.5\%          \\
ResNet-152 + IDT                                             & 92.0\%          \\
\textbf{P3D ResNet} + IDT                                & \textbf{93.7\%}          \\ \hline
\end{tabular}
\label{tab:ucf101}
\vspace{-0.25in}
\end{table}

\textbf{Comparison with the state-of-the-art.} We first compare with several state-of-the-art techniques in the context of video action recognition on three splits of UCF101 and ActivityNet validation set. The performance comparisons are summarized in Table \ref{tab:ucf101} and \ref{tab:activitynet}, respectively. We briefly group the approaches on UCF101 into three categories: end-to-end CNN architectures which are fine-tuned on UCF101 in the upper rows, CNN-based video representation extractors with linear SVM classifier in the middle rows and approaches fused with IDT in the bottom rows. It is worth noting that most recent end-to-end CNN architectures on UCF101 often employ and fuse two or multiple types of inputs, e.g., frame, optical flow or even audio. Hence, the performances by exploiting only video frames and late fusing the scores on two inputs of video frames plus optical flow are both reported. As shown in Table \ref{tab:ucf101}, the accuracy of P3D ResNet can achieve 88.6\%, making the absolute improvement over the best competitor TSN on the only frame input and ResNet-152 in the first and second category by 2.9\% and 5.1\%, respectively. Compared to \cite{yue2015beyond} which operates LSTM over high-level representations of frames to explore temporal information, P3D ResNet is benefited from the temporal connections throughout the whole architecture and outperforms \cite{yue2015beyond}. P3D ResNet with only frame input is still superior to \cite{simonyan2014two,sun2015human,yue2015beyond} when fusing the results on the inputs of both frame and optical flow. The results also consistently indicate that fusing two kinds of inputs (performances in brackets) leads to apparent improvement compared to only using video frames. This motivates us to learn P3D ResNet architecture with other types of inputs in our future works. Furthermore, P3D ResNet utilizing 2D spatial convolutions plus 1D temporal convolutions exhibits significantly better performance than C3D which directly uses 3D spatio-temporal convolutions. By combining with IDT \cite{wang2013action} which are hand-crafted features, the performance will boost up to 93.7\%. In addition, by performing the recent state-of-the-art encoding method \cite{qiu2017deep} on the activations of res5c layer in P3D ResNet, the accuracy can achieve 90.5\%, making the improvement over the representation from pool5 layer in P3D ResNet by 1.9\%.

\begin{table}
\centering
\small
\caption{\small Performance comparisons in terms of Top-1\&Top-3 classification accuracy, and mean AP on ActivityNet validation set. A linear SVM classifier is learnt on each feature.}
\begin{tabular}{l|c|c|c} \hline
\textbf{Method}                                             & \textbf{Top-1} & \textbf{Top-3} & \textbf{MAP} \\ \hline
IDT \cite{wang2013action}                                   & 64.70\%        & 77.98\%        & 68.69\% \\
C3D \cite{tran2015learning}                                 & 65.80\%        & 81.16\%        & 67.68\% \\
VGG\_19 \cite{Simonyan:ICLR15}                              & 66.59\%        & 82.70\%        & 70.22\% \\
ResNet-152 \cite{he2015deep}                                & 71.43\%        & 86.45\%        & 76.56\% \\ \hline
\textbf{P3D ResNet}                                         & \textbf{75.12\%}        & \textbf{87.71\%}        & \textbf{78.86\%} \\ \hline
\end{tabular}
\label{tab:activitynet}
\vspace{-0.15in}
\end{table}

The results across different evaluation metrics constantly indicate that video representation produced by our P3D ResNet attains a performance boost against baselines on ActivityNet validation set, as shown in Table \ref{tab:activitynet}. Specifically, P3D ResNet outperforms IDT, C3D, VGG\_19 and ResNet-152 by 10.4\%, 9.3\%, 8.5\% and 3.7\% in terms of Top-1 accuracy, respectively. There is also a large performance gap between C3D and ResNet-152. This is mainly due to data shift that the categories in ActivityNet are mostly human activities in daily life, which are quite different from those sport-related data in Sports-1M benchmark, resulting in not satisfying performance by C3D learnt purely on Sports-1M data. Instead, ResNet-152 trained on ImageNet image data is found to be more helpful in this case. P3D ResNet which pre-trains 2D spatial convolutions on image data and learns 1D temporal convolutions on video data fully leverages the knowledge from two domains, successfully boosting up the performance.

\begin{table}
\centering
\small
\caption{\small Action similarity labeling performances on ASLAN benchmark. STIP: Space-Time Interest Points; MIP: Motion Interchange Patterns; FV: Fisher Vector.}
\begin{tabular}{l|c|c|c} \hline
\textbf{Method}                                             & \textbf{Model} & \textbf{Accuracy} & \textbf{AUC} \\ \hline
STIP \cite{KG:PAMI12}                                & linear         & 60.9\%        & 65.3\% \\
MIP \cite{kliper2012motion}                                 & metric         & 65.5\%        & 71.9\% \\
IDT+FV \cite{peng2014large}                                 & metric         & 68.7\%        & 75.4\% \\
C3D \cite{tran2015learning}                                 & linear         & 78.3\%        & 86.5\% \\
ResNet-152 \cite{he2015deep}                                 & linear         & 70.4\%        & 77.4\% \\ \hline
\textbf{P3D ResNet}                                      & linear         & \textbf{80.8\%}        & \textbf{87.9\%} \\
\hline
\end{tabular}
\label{tab:aslan}
\vspace{-0.22in}
\end{table}

The second task is action similarity labeling challenge, which is to answer a binary question of ``does a pair of videos present the same action?" Following the experimental settings in \cite{KG:PAMI12,tran2015learning}, we extract the outputs of four layers in P3D ResNet, i.e., prob, pool5, res5c and res4b35 layer as four types of representation for each 16-frame video clip. The video-level representation is then obtained by averaging all clip-level representations. Given each video pair, we calculate 12 different similarities on each type of video representation and thus generate a 48-dimensional vector for each pair. An L2 normalization is implemented on the 48-d vector and a binary classifier is trained by using linear SVM. The performance comparisons on ASLAN are shown in Table \ref{tab:aslan}. Overall, P3D ResNet performs consistently better than both hand-crafted features and CNN-based representations across the performance metric of accuracy and area under ROC curve (AUC). In general, CNN-based representations exhibits better accuracy than hand-crafted features. Unlike the observations on action recognition task, C3D significantly outperforms ResNet-152 on the scenario of action similarity labeling. We speculate that this may be the result of difficulty in interpreting the similarity between videos based on the ResNet-152 model learnt purely on image domain. In contrast, the video representation extracted by C3D which is trained on video data potentially has higher capability to distinguish between videos. At this point, improvements are also observed in P3D ResNet. This again indicates that P3D ResNet is endowed with the advantages of both C3D and ResNet-152 by pre-training 2D spatial convolutions on image data and learning 1D temporal connections on video data.

The third experiment was conducted on scene recognition. Table \ref{tab:scene} shows the accuracy of different methods. P3D ResNet outperforms the state-of-the-art hand-crafted features \cite{feichtenhofer2014bags} by 16.9\% and 3.3\% on Dynamic Scene and YUPENN benchmark, respectively. Compared to C3D and ResNet-152, P3D ResNet makes the absolute improvements by 1.4\% and 0.3\% on YUPENN, respectively.

%In particular, the accuracy of P3D ResNet will boost up to 99.5\% on this benchmark.

\begin{table}
\centering
\small
\caption{\small The accuracy performance of scene recognition on Dynamic Scene and YUPENN sets.}
\begin{tabular}{l|c|c} \hline
\textbf{Method}                                          & \textbf{Dynamic Scene} & \textbf{YUPENN} \\ \hline
\cite{Der:CVPR12}                               & 43.1\%                 & 80.7\% \\
\cite{feichtenhofer2014bags}                             & 77.7\%                 & 96.2\% \\
C3D \cite{tran2015learning}                              & 87.7\%                 & 98.1\% \\
ResNet-152 \cite{he2015deep}                              & 93.6\%                 & 99.2\% \\ \hline
\textbf{P3D ResNet}                                   & \textbf{94.6\%}                 & \textbf{99.5\%} \\
\hline
\end{tabular}
\label{tab:scene}
\vspace{-0.15in}
\end{table}

\begin{figure}[!tb]
   \centering {\includegraphics[width=0.35\textwidth]{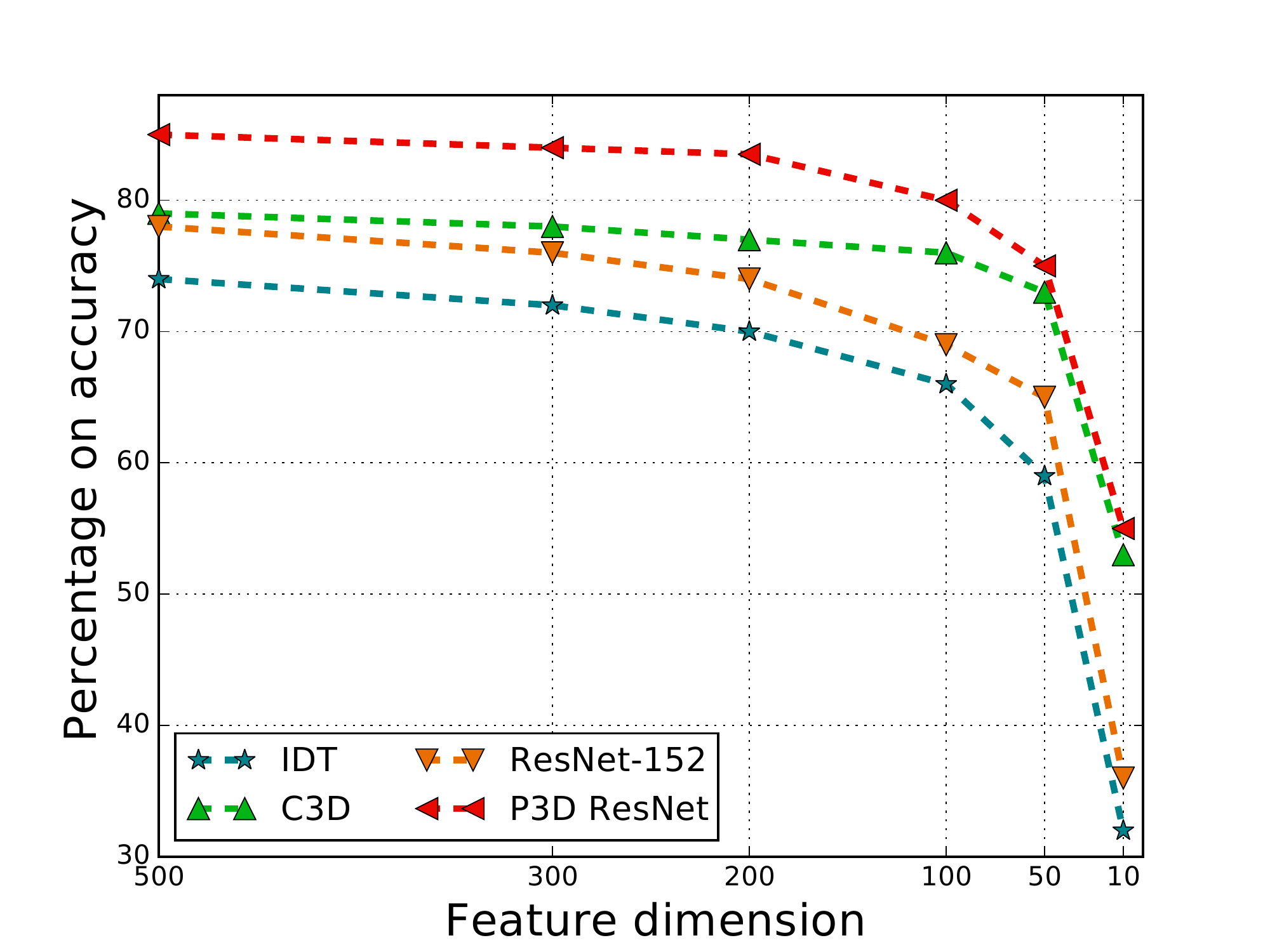}}
   \caption{\small The accuracy of video representation learnt by different architectures with different dimensions. The performances reported in this figure are on UCF101 (3 splits).}
   \label{fig:fig5}
   \vspace{-0.2in}
\end{figure}

\textbf{The effect of representation dimension.} Figure \ref{fig:fig5} compares the accuracy of video representation with different dimensions on UCF101 by performing Principal Components Analysis on the original features of IDT, ResNet-152, C3D and P3D ResNet. Overall, video representation learnt by P3D ResNet consistently outperforms others at each dimension from 500 to 10. In general, higher dimensional representation provide better accuracy. An interesting observation is that the performance of ResNet-152 decreases more sharply than that of C3D and P3D ResNet when reducing the representation dimension. This somewhat reveals the weakness of ResNet-152 in generating video representation, which is originated from domain gap that ResNet-152 is learnt purely on image data and may degrade the representational capability on videos especially when the feature dimension is very low. P3D ResNet, in comparison, is benefited from the exploration of knowledge from both image and video domain, making the learnt video representation more robust to the change of dimension.

\begin{figure}[!tb]
   \centering
   \subfigure[ResNet-152]{
     \label{fig:fig8:a}
     \includegraphics[width=0.2\textwidth]{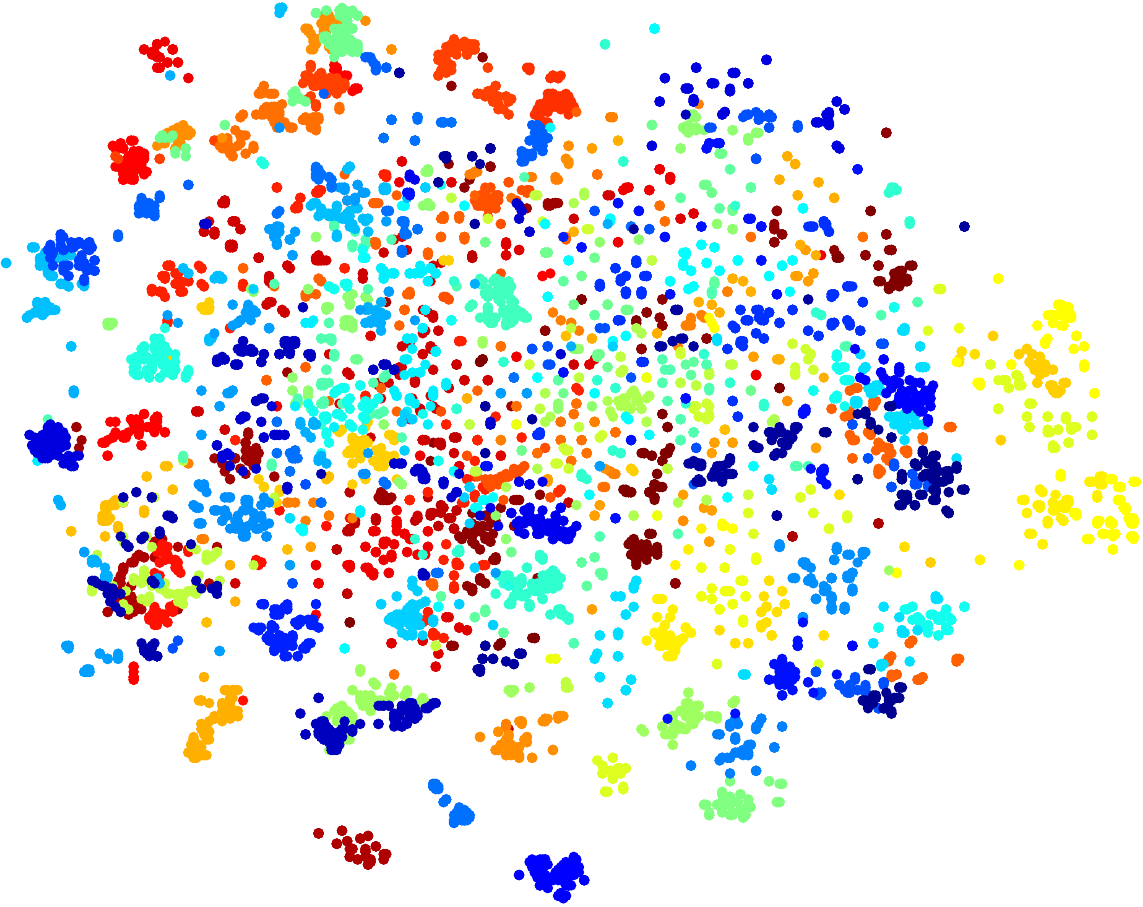}}
   \subfigure[P3D ResNet]{
     \label{fig:fig8:b}
     \includegraphics[width=0.2\textwidth]{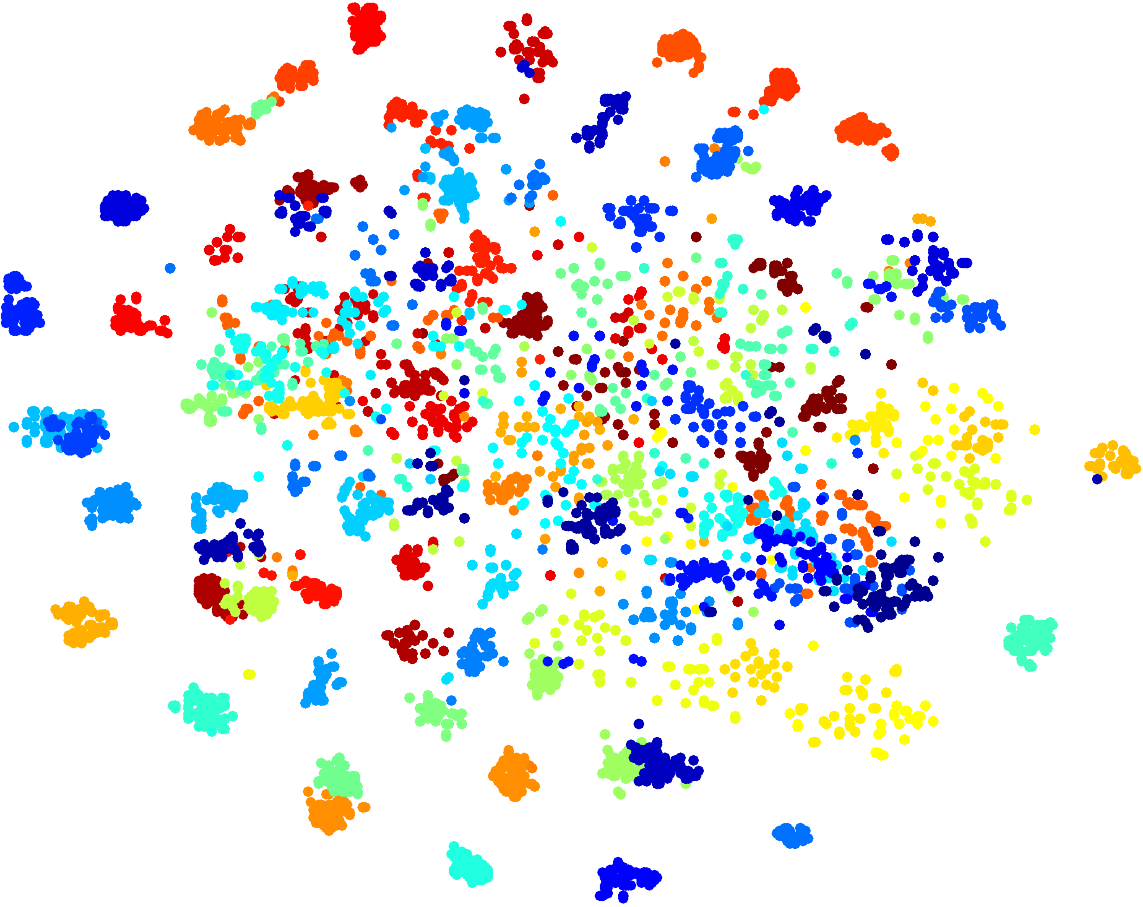}}
   \caption{\small Video representation embedding visualizations of ResNet-152 and P3D ResNet on UCF101 using t-SNE \cite{maaten:JMLR08}. Each video is visualized as one point and colors denote different actions.}
   \label{fig:fig8}
   \vspace{-0.22in}
\end{figure}

\textbf{Video representation embedding visualization.} Figure \ref{fig:fig8} further shows the t-SNE \cite{maaten:JMLR08} visualization of embedding of video representation learnt by ResNet-152 and P3D ResNet. Specifically, we randomly select 10K videos from UCF101 and the video-level representation is then projected into 2-dimensional space using t-SNE. It is clear that video representations by P3D ResNet are better semantically separated than those of ResNet-152.

\section{Conclusion}
We have presented Pseudo-3D Residual Net (P3D ResNet) architecture which aims to learn spatio-temporal video representation in deep networks. Particularly, we study the problem of simplifying 3D convolutions with 2D filters on spatial dimension plus 1D temporal connections. To verify our claim, we have devised variants of bottleneck building blocks for combining the 2D spatial and 1D temporal convolutions, and integrated them into a residual learning framework at different placements for structural diversity purpose. The P3D ResNet architecture learnt on Sports-1M dataset validate our proposal and analysis. Experiments conducted on five datasets in the context of video action recognition, action similarity labeling and scene recognition also demonstrate the effectiveness and generalization of the spatio-temporal video representation produced by our P3D ResNet. Performance improvements are clearly observed when comparing to other feature learning techniques.

Our future works are as follows. First, attention mechanism will be incorporated into our P3D ResNet for further enhancing representation learning. Second, an elaborated study will be conducted on how the performance of P3D ResNet is affected when increasing the frames in each video clip in the training. Third, we will extend P3D ResNet learning to other types of inputs, e.g., optical flow or audio.

{\small
\bibliographystyle{ieee}
\bibliography{egbib}
}

\end{document}